\documentclass[conference]{IEEEtran}
\usepackage{cite}
\usepackage{amsmath,amssymb,amsfonts}
\usepackage{algorithmic}
\usepackage{graphicx}
\usepackage{textcomp}
\usepackage{xcolor}

\usepackage{subcaption}
\usepackage{diagbox}
\usepackage[bookmarks=true,breaklinks=true,letterpaper=true,colorlinks,linkcolor=black,citecolor=magenta,urlcolor=blue]{hyperref}

\def\BibTeX{{\rm B\kern-.05em{\sc i\kern-.025em b}\kern-.08em
    T\kern-.1667em\lower.7ex\hbox{E}\kern-.125emX}}
\begin{document}

% \title{Conference Paper Title*\\
\newcommand{\SYSNAME}{ClusterKV}
\newcommand{\fix}[1]{{\textcolor{red}{#1}}}

\title{\SYSNAME{}: Manipulating LLM KV Cache in Semantic Space for Recallable Compression}

\author{
    \IEEEauthorblockN{Guangda Liu, Chengwei Li, Jieru Zhao\textsuperscript{$\dagger$}, Chenqi Zhang, Minyi Guo}
    \IEEEauthorblockA{School of Computer Science, Shanghai Jiao Tong University\\
    % \{gd.liu, zhao-jieru\}@sjtu.edu.cn 
    \textsuperscript{$\dagger$} Corresponding author: zhao-jieru@sjtu.edu.cn
    % \{gd.liu, exia, zhao-jieru, zhangchenqi123\}@sjtu.edu.cn, guo-my@cs.sjtu.edu.cn
    }
}

% \author{
% \IEEEauthorblockN{Author 1\IEEEauthorrefmark{1}, Author 2\IEEEauthorrefmark{1}, Author 3\IEEEauthorrefmark{2}}
% \IEEEauthorblockA{\IEEEauthorrefmark{1}Affiliation 1, City, Country\\
% \IEEEauthorrefmark{2}Affiliation 2, City, Country}
% }

\maketitle

% \begingroup\renewcommand\thefootnote{*}
% \footnotetext{Corresponding author: Jieru Zhao (zhao-jieru@sjtu.edu.cn).}
% \endgroup

\begin{abstract}
Large Language Models (LLMs) have been widely deployed in a variety of applications, and the context length is rapidly increasing to handle tasks such as long-document QA and complex logical reasoning.
However, long context poses significant challenges for inference efficiency, including high memory costs of key-value (KV) cache and increased latency due to extensive memory accesses.
Recent works have proposed compressing KV cache to approximate computation, but these methods either evict tokens permanently, never recalling them for later inference, or recall previous tokens at the granularity of pages divided by textual positions. Both approaches degrade the model accuracy and output quality.
To achieve efficient and accurate recallable KV cache compression, we introduce \SYSNAME{}, which recalls tokens at the granularity of semantic clusters. We design and implement efficient algorithms and systems for clustering, selection, indexing and caching. 
Experiment results show that \SYSNAME{} attains negligible accuracy loss across various tasks with 32k context lengths, using only a 1k to 2k KV cache budget, and achieves up to a 2$\times$ speedup in latency and a 2.5$\times$ improvement in decoding throughput.
Compared to SoTA recallable KV compression methods, \SYSNAME{} demonstrates higher model accuracy and output quality, while maintaining or exceeding inference efficiency. Our code is available at \url{https://github.com/sjtu-zhao-lab/ClusterKV}.
\end{abstract}

% \begin{IEEEkeywords}
% component, formatting, style, styling, insert
% \end{IEEEkeywords}

\vspace{-0.2cm}
\section{Introduction}
Large Language Models (LLMs) have been widely deployed in a variety of applications, such as natural language understanding, coding copilots and chatbots.
As LLMs are used for more complex tasks, the need for long context length is rapidly increasing to handle tasks such as long-document QA, repository-level code understanding, and complex logical reasoning \cite{longbench, swebench, gpt-o1}.
This drives LLMs to support larger context windows, expanding from the original 4k to 32k \cite{llama2,longchat2023}, 128k~\cite{gpt-4o} and even up to 1M \cite{glm2024}.
While models can support larger context window with continual training or extrapolation~\cite{longchat2023}, inference with long contexts poses significant challenges on efficiency.
As the size of key-value (KV) cache increases linearly with the context length, it can exceed the capacity of GPU memory, leading to inference failure or extremely high latency.
Since the autoregressive decoding stage is typically the performance bottleneck of LLM inference and is memory-bound \cite{splitwise}, accessing the KV cache of all previous tokens results in increased latency for long-context inference.

\begin{figure}[t]
	\centering
	\includegraphics[width=1\linewidth]{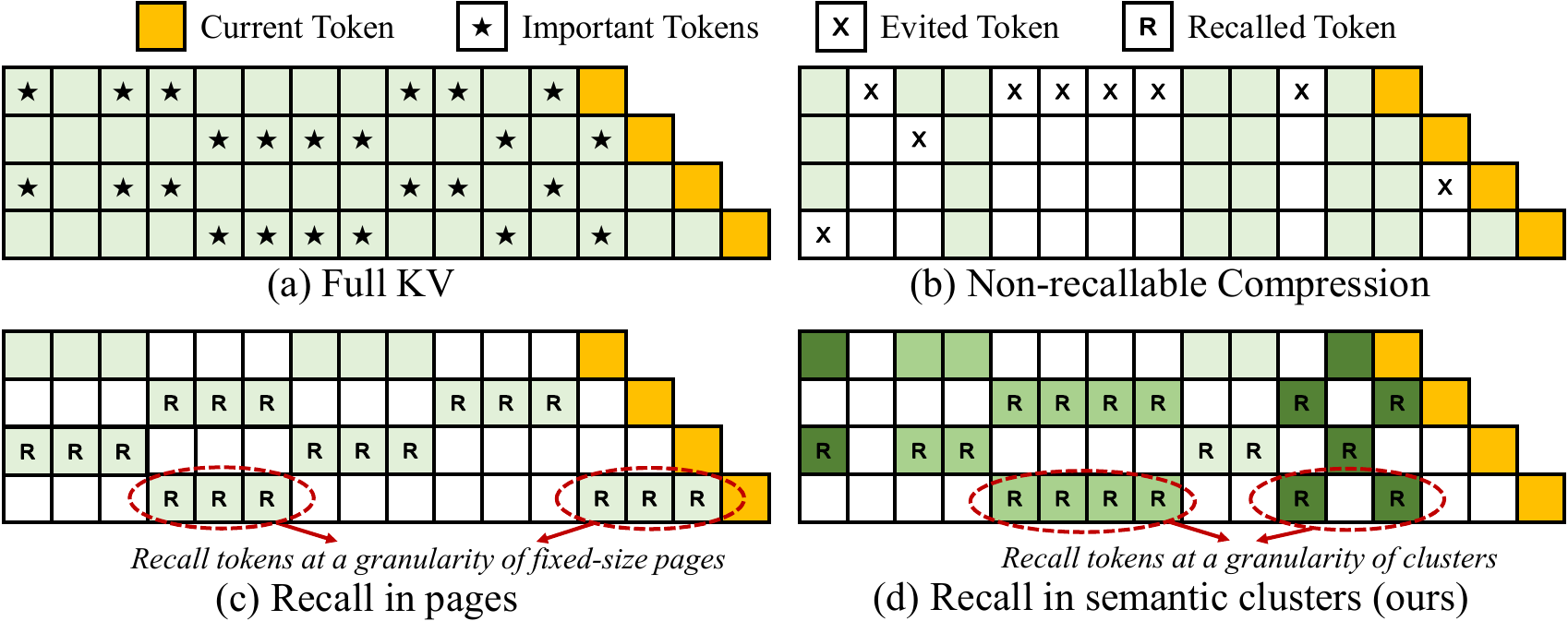}
	\caption{Comparison of KV compression methods. Green boxes represent tokens selected for attention computation.
 % Full KV, Not recall, Recall at block level, Recall in semantic space
 }
	\label{fig:intro-cmp}
    \vspace{-0.3cm}
\end{figure}

To mitigate the issues, recent works 
% leverage the sparsity of attention computation and 
compress KV cache by selecting a subset of tokens and utilizing their keys/values
% tokens 
to approximate attention computation.
The tokens are typically selected with fixed patterns \cite{streamingllm} or based on their attention weights \cite{h2o,snapkv,keyformer,alisa,inf-mllm}, which represent each token's importance in attention computation.
% However, 
In most compression methods, unimportant tokens are evicted permanently; once evicted, they are never recalled in subsequent decoding steps, as shown in Fig.~\ref{fig:intro-cmp}b. 
% This is because recalling evicted tokens requires computing attention weights over all previous tokens, incurring unacceptable selection overhead.
However, the token importance is dynamically changing during decoding: tokens initially considered unimportant and evicted may become important at later steps and vice versa. 
Non-recallable KV cache compression greedily evicts tokens based on positions or attention weights of tokens at the current decoding step, failing to capture the dynamic feature.  
% Non-recallable KV cache compression can degrade model accuracy and output quality, as the importance of KV tokens is dynamic. 
This can degrade model accuracy and output quality.
% while non-recallable methods fail to capture tokens with dynamic importance.

\begin{figure}[t]
	\centering
	\includegraphics[width=1\linewidth]{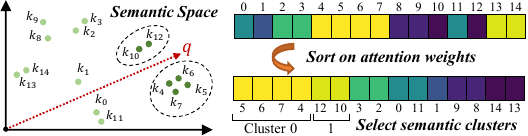}
	\caption{Semantic space and attention weights of tokens in the last step of Fig. \ref{fig:intro-cmp}d. Lighter boxes indicate larger weights.}
	\label{fig:intro-cluster}
    \vspace{-0.5cm}
\end{figure}

While making tokens recallable 
% to capture those tokens with dynamic importance 
is promising, it incurs prohibitively high cost due to the computation of attention weights over all previous tokens. 
% \fix{add sentences to connect the previous and this paragraph.}
% To enable recallable KV cache compression at an acceptable cost, 
To reduce the cost, as illustrated in Fig. \ref{fig:intro-cmp}c, Quest proposed recalling at the granularity of pages \cite{quest}.
A page consists of $page\_size$ consecutive tokens, allowing Quest to reduce the selection overhead by a factor of $1/page\_size$.
However, since pages are divided simply by textual positions of tokens, internal fragmentation becomes an issue: a recalled page may contain unimportant tokens, wasting budget that could be allocated to truly important tokens. 

To achieve efficient and accurate KV cache compression, we introduce \SYSNAME{}, which recalls tokens at the granularity of token clusters in the semantic space, referred to as \textit{semantic clusters}.
As illustrated in Fig. \ref{fig:intro-cluster}, tokens that are close in their semantic space or key vector space exhibit similar attention weights.
Therefore, as shown in Fig. \ref{fig:intro-cmp}d, \SYSNAME{} recalls tokens at the granularity of semantic clusters to ensure more precise token recall.
Moreover, \SYSNAME{} incorporates specialized system designs and optimized kernels to minimize recall overheads.
Experimental results demonstrate that \SYSNAME{} attains negligible accuracy loss across various tasks with context lengths up to 32k, using only a 1k to 2k KV cache budget. 
It also delivers up to a 2$\times$ speedup in latency and a 2.5$\times$ improvement in decoding throughput.
Compared to state-of-the-art recallable KV compression methods, \SYSNAME{} achieves significantly higher model accuracy and output quality, while maintaining or surpassing inference efficiency.
\vspace{-0.1cm}
\section{Background and Motivation}
\subsection{LLM Inference and KV Cache}
% Transformer-based 
LLMs encompass multiple Transformer layers, each containing a multi-head attention (MHA) module and a feed-forward network (FFN) with residual connections and normalization operations \cite{transformer}.
In MHA, the input tensor is linearly projected into query, key and value tensors ($Q, K, V \in \mathbb{R}^{N\times d}$) for each head, where $N$ is the input length and $d$ represents the number of channels or hidden dimensions per head.
The MHA output is defined as $softmax(\frac{QK^T}{\sqrt{d}})V$, with outputs of all heads concatenated for subsequent FFN and normalization.

For generative inference, LLMs produce tokens in an autoregressive manner, appending each generated token to the input to generate the next token. 
To avoid the recomputation of $K$ and $V$ of previous tokens, these tensors are stored in memory for reuse, which is known as \textit{KV cache}. 
The LLM inference includes two stages: prefill and decoding.
% can be split to two stages: prefill and decoding. 
The prefill stage processes the entire input sequence, computes KV cache and generates the first output token.
During decoding, the query vector $q$ of the latest generated token and $K, V$ of previous tokens are used to compute attention for generating the next token, formulated as $softmax(\frac{qK^T}{\sqrt{d}})V$, 
where $q\in \mathbb{R}^{1\times d}$, $K, V \in \mathbb{R}^{L\times d}$ and $L$ is the context length of previous tokens.

\vspace{-0.1cm}
\subsection{Long Context Inference and KV Cache Compression}
\label{sec:background-compression}
Long context is an emerging trend of LLM inference,
% On the one hand, Tasks 
such as long-document QA \cite{longbench, infinibench}, repository-level code understanding \cite{swebench} and complex logical reasoning \cite{gpt-o1}. 
% require extremely large input length.
% On the other hand, 
Moreover, supported context windows of LLMs are extending rapidly, from 4k tokens of GPT3.5 and Llama-2 to 32k \cite{longchat2023}, 128k~\cite{gpt-4o} and even 1M tokens \cite{glm2024}.
However, long-context inference incurs significant memory and computation cost. The size of KV cache and complexity of attention computation during decoding increase linearly with context length, resulting in low inference efficiency or even inference failures.
% For example?

% To address these performance issues, most of LLMs adapt MHA to grouped query attention (GQA) \cite{gqa}, where multiple attention heads share the same KV cache of a KV head. Although GQA is effective \cite{llama3,razor-attn,duo-attn}, it introduces additional training costs and does not mitigate the linear growth in cost as the context length increases.

Recent works reveal the sparsity of attention computation, i.e., only a small subset of tokens contributes to most of attention outputs \cite{h2o, alisa}. 
This observation enables KV cache compression by selecting a subset of tokens to approximate attention computation, which is formulated as $softmax(\frac{qK_S^T}{\sqrt{d}})V_S$,
% \begin{equation}
% 	\label{eq:sel-attn}
% 	\Hat{Attn}(q, K, V) = softmax(\frac{qK_S^T}{\sqrt{d}})V_S
% \end{equation}
where $K_S, V_S \in \mathbb{R}^{B\times d}$ represents the keys and values of selected tokens, and $B$ is the budget size of KV cache after compression.
By setting a fixed budget, the KV cache size and decoding cost remain stable, regardless of the context length. 
% Moreover, this technique does not require additional training.
Typically, $K_S$ and $V_S$ are selected based on attention weights ($qK^T$), as tokens with higher attention weights contributes more to the attention computation \cite{h2o,snapkv,keyformer}.

% \fix{can we make the following examples concise? introduce examples in general and add a series of citations together?}
% For example, H2O \cite{h2o} computes accumulated attention scores of recent several steps, and selects KV tokens with largest accumulated scores, i.e. $qK^T$. 
% And SnapKV \cite{snapkv} computes attention weights of a context window, averaging over the window and selecting tokens with largest average attention weights.

% \footnote{In this work, we refer to the results of $qK^T$ and $softmax(qK^T)$ as attention weights and attention scores, respectively.}. 
\vspace{-0.2cm}
\subsection{Related Work and Motivation}
\vspace{-0.1cm}

\begin{figure}[t]
    \centering
    \begin{subfigure}[b]{1\linewidth}
        \centering
        \includegraphics[width=\textwidth]{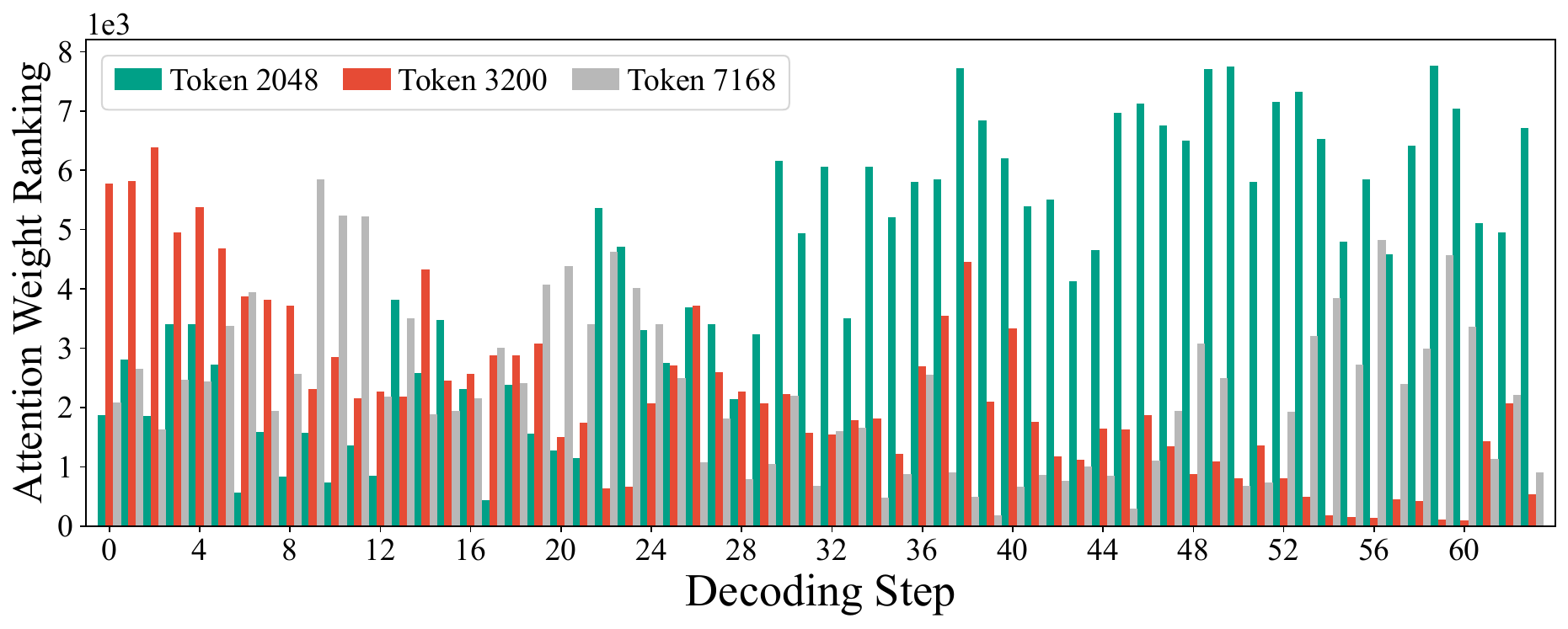}
        \caption{}
        \label{fig:motivation-dyn}
    \end{subfigure}\vfill
    \begin{subfigure}[b]{1\linewidth}
        \centering
        \includegraphics[width=\textwidth]{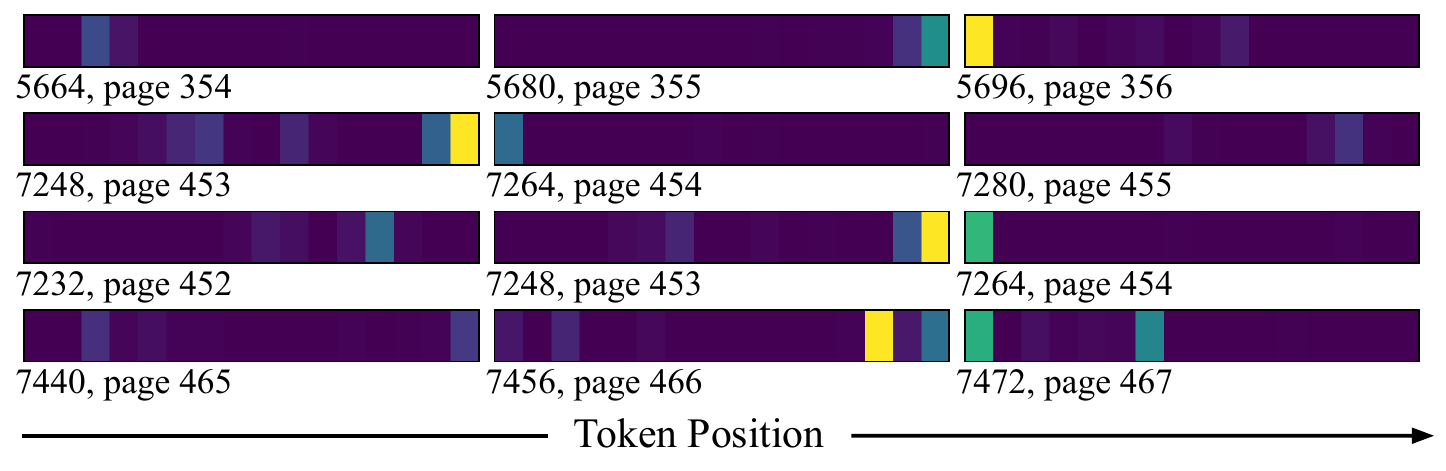}
        \caption{}
        \label{fig:motivation-page}
    \end{subfigure}
    \caption{(a) Variation in token importance across decoding steps with a context length of 8192. (b) Internal fragmentation of important tokens at the granularity of pages ($page\_size=16$).}
    \label{fig:motivation}
    \vspace{-0.5cm}
\end{figure}

% \noindent\textbf{Dynamism of Token Importance.}
\textbf{KV cache compression should be recallable.}
While selecting a subset of KV cache for computation ensures inference efficiency, computing all the attention weights for selection introduces substantial overhead.
% brings additional overheads, particularly as FlashAttention is now adopted by most inference frameworks and attention weights are not explicitly computed \cite{flash-attn2}.
Therefore, 
% to reduce the selection overheads, 
most existing works only compute attention weights over tokens that have been selected, rather than for all previous tokens \cite{h2o,snapkv,keyformer}. 
In this way, keys and values of tokens not selected at one decoding step are permanently evicted from the KV cache and never recalled in later inference steps, 
% We refer this kind of method as \textit{non-recallable} KV compression, 
as shown in Fig. \ref{fig:intro-cmp}b.

However, we observe that the token importance changes dynamically during inference. 
Token that are unimportant with low attention weights at one decoding step can become important in later steps, and vice versa. 
% This phenomenon is demonstrated in Fig. 
Figure \ref{fig:motivation-dyn} shows changes in attention weight rankings during 64 decoding steps of Llama-3-8B.
For instance, token 3200 
% starts with a low ranking and 
is initially unimportant but becomes crucial after 20 steps, while the opposite occurs for token 2048. And importance of all tokens can fluctuate throughout inference, as seen with token 7168.
Therefore, non-recallable compression inevitably overlooks some tokens with dynamically changing importance, leading to reduced model accuracy and a decline in output quality.
% model accuracy drops and lower quality of model outputs.

% \noindent\textbf{Defects of blockwise selection.}
\textbf{Defects of existing recallable KV compression methods.}
% While making KV cache compression recallable is essential for maintaining model accuracy, 
However, achieving recallable compression by computing attention weights with all previous tokens incurs an unacceptable cost of $O(Ld)$, which is comparable to attention computation with full KV cache and negates the benefits of compression.
To reduce the cost, InfiniGen \cite{infinigen} reduces the dimensions of $q$ and $K$ by generating partial query and key weights offline using singular value decomposition, and projecting hidden states with these partial weights. 
However, it requires generating and storing partial keys in addition to original keys, and the selection cost still scales linearly with the context length $L$.

% On the other hand, 
Quest \cite{quest} selects tokens at the granularity of pages, which consist of several consecutive tokens, as depicted in Fig. \ref{fig:intro-cmp}c. 
To estimate importance of a page, it uses the per-channel maximal keys of all tokens within a page to compute attention weights. This approach reduces the selection cost to $O(Ld/{page\_size})$ as only one attention weight is needed for every $page\_size$ tokens.
However, 
recalling pages which are simply divided by token positions leads to internal fragmentation of important tokens. 
A selected page may contain only a few important tokens, while including unimportant ones for attention computation and wasting the KV cache budget. 
We illustrate this issue using attention heatmaps of Llama3-8B with an 8k context length in Fig. \ref{fig:motivation-page}, where lighter cells represent more important tokens with higher attention weights. 
As shown, each page of 16 tokens contains only one or two important tokens, and in some cases, two pages or 32 tokens are required to include two consecutive important tokens.

% Based on the above observations, 
% To solve these issues, we introduce \SYSNAME{}, achieving efficient recallable KV cache compression by selecting in the granularity of semantic clusters rather than pages divided by textually positions. 
% \fix{Fig. XX, may be merged with the new figure in intro if the space is not sufficient?}

\section{Algorithm Design}
\subsection{Problem Formulation and Design Rationale}
\textbf{Problem Formulation.} For approximated attention computation with selected tokens formulated in Sec. \ref{sec:background-compression}, let $K_S$ to be $(k_{i_1}, k_{i_1}, ..., k_{i_B})^T$, where ${I_T}=\{i_1, i_2, ..., i_B\}$ denotes the indices of selected tokens.
Our goal is to select tokens which contribute most to attention weights, thereby approximating the original computation as closely as possible. 
Specifically, ${I_T}$ is supposed to be $\mathrm{arg\,max}\sum_{i\in {I_T}} qk^T_{i}$, which corresponds to selecting tokens with top-$B$ largest attention weights.

\textbf{Design Rationale.} 
% As previously mentioned, computing $qk^T_i$ over all previous tokens for selection incurs a cost comparable to that of the attention computation with full KV cache, which is therefore unacceptable.
% To mitigate the KV selection cost, we propose our clustering approach based on the intuition: 
Our observation is that \textit{tokens which are close in semantic space tend to have similar attention weights for a given $q$}.
Therefore, we propose KV selection at the granularity of semantic clusters. 
We first apply clustering to tokens in the semantic space. Then, we compute attention weights only with respect to the cluster representations, rather than individual tokens, and select clusters with the largest attention weights.
Since the number of clusters is typically an order of magnitude smaller than the number of tokens, cluster-based selection significantly reduces recall overhead.

% \vspace{-0.1cm}
\subsection{Clustering in the Semantic Space}
\label{sec:alg_cluster}
% \vspace{-0.1cm}
\textbf{Semantic Distance.} 
% Specifically, 
As for a given $q$, attention weights are associated only with the key tensors, we measure the semantic distance between tokens by calculating the distance between corresponding key vectors.
Furthermore, we find that \textit{cosine similarity} is more suitable than L2 or inner product distances. This is due to the presence of outlier channels with large magnitudes in key vectors \cite{kivi}, which can cause drastic changes in L2 or inner product distances. 
% We demonstrate the impact of different distance measures in Section \ref{sec:eval-acc}.
Thus, we define distance between token $i$ and token $j$ in the semantic space as $\mathcal{D}(i, j)=1-\frac{\langle k_i, k_j \rangle}{|k_i|\cdot|k_j|}$, where distance is smaller for vectors with larger cosine similarity.

\begin{figure}[t]
	\centering
	\includegraphics[width=1\linewidth]{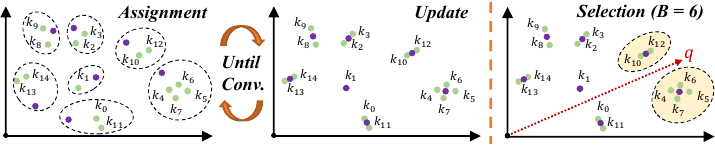}
	\caption{Clustering and selection process. Green dots represent key vectors, and purple dots represent centroids of clusters.}
	\label{fig:algo}
    \vspace{-0.5cm}
\end{figure}

\textbf{Clustering.} We apply a simple K-means algorithm for clustering over key vectors as shown in Fig. \ref{fig:algo} \cite{kmeans}.
We first randomly sample key vectors as the initial centroids. Subsequently, we alternatively perform the assignment and update steps until convergence. 
In the assignment step, each key vector is assigned to the nearest centroid and given a corresponding \textit{cluster label} based on the distance $\mathcal{D}$, i.e., assigned to the centroid with the maximum cosine similarity.
Then in the update step, the mean of keys assigned to the same centroid are used as the new centroid.
The algorithm converges when the assignment no longer changes, and keys assigned to the same centroid form a \textit{semantic cluster}, with the centroid as the cluster representation.

During LLM inference, we first apply clustering to the key vectors of prompt tokens \textit{after the prefill stage}.
However, we note an exception for the initial tokens, referred to as attention sinks \cite{streamingllm}. These tokens typically appear as outliers in the clustering process, as they are distant from other tokens in the semantic space. 
Therefore, we always retain the first 16 tokens and apply clustering to the subsequent tokens, generating $C_0$ centroids. 
We set $C_0=\frac{L}{80}$, as our experiments indicate that for a 32k context, using 400 clusters achieves a balance between efficiency and accuracy.
For tokens generated \textit{in the decoding stage}, clustering is applied every $m$ decoding steps to the key vectors of $m$ generated tokens, creating $C_+$ new centroids.
Since clustering keys of generated tokens together with those from the prefill stage can incur significant overhead, we instead apply clustering \textit{within} the generated tokens only.
To amortize the cost, we set $C_+$ and $m$ to 4 and 320, respectively.
% \fix{neighboring tokens tend to have close semantic distance? but this "neighboring" is similar to Quest, is there a better description?}, 

% we apply clustering to their key vectors individually, rather than clustering them together with key vectors from the prefill stage. 
% This is because the cost of one assignment step for clustering all tokens will be $O(CLd)$, which has exceeded the cost of decoding with full KV cache, $O(Ld)$.
% To accumulate key vectors for clustering and amotize the cost, clustering is applied every 256 generated tokens during the decoding stage.

\vspace{-0.1cm}
\subsection{Selection at the Granularity of Semantic Clusters}
\label{sec:algo-sel}
\vspace{-0.1cm}
\textbf{Selection.} We denote cluster centroids as $\mu_1, \mu_2, ..., \mu_C \in \mathbb{R}^d$. 
To select important tokens for a given query $q$, we sort those centroids based on their attention weights, i.e., $q\mu^T_i$, in descending order.
While keys are clustered using cosine similarity distance, the distance between query vector and centroids is measured with inner product, as it better aligns with attention weight computation.
% The indices of sorted centroids as $I_C$ is denoted as $I_C$.
% Based on our intuition, 
Intuitively, keys assigned to clusters with centroids that have larger attention weights tend to have larger attention weights for the given $q$. 
Therefore, we retrieve the sorted centroids and collect KV of tokens from the corresponding clusters, until the top-$B$ most important tokens are selected, as shown in Fig. \ref{fig:algo}.

\vspace{-0.1cm}
\subsection{Efficiency Concerns}
\vspace{-0.1cm}
Cluster-based selection
% at the granularity of semantic clusters 
avoids the internal fragmentation of important tokens and can achieve higher accuracy compared to page-based selection. However, it incurs efficiency concerns.

\textbf{Concern 1.} For clustering, the computational cost is $O(n_iCLd)$ where $n_i$ is the number of iterations until convergence and $C$ is the number of clusters, which is higher than the cost of obtaining page representations, such as per-channel maximal key vectors in Quest \cite{quest}, which is $O(Ld)$. 

\textbf{Concern 2.}
For selection in page-based methods (Fig. \ref{fig:intro-cmp}c), since page size is fixed and tokens within a page are consecutive, the number of needed pages can be easily computed as $\frac{B}{page\_size}$, and the indices of selected tokens can be directly derived from the indices of selected pages.
However, for semantic clusters, the sizes can vary, and the token positions within a cluster are dynamic and discontinuous. 
% As a result, 
Therefore, the number of clusters needed and the indices of selected tokens cannot be easily determined in the same way as for pages.

% These complexities poses challenges to inference efficiency when selecting at the granularity of semantic clusters.
%  In the following section, we describe the system design and implementation of \SYSNAME{}, which addresses these challenges and maintains the inference efficiency.
% To address these concerns and maintain high efficiency, we introduce our system design of \SYSNAME{} in Sec. \ref{sec:system}.
Our system design addresses both concerns in Sec. \ref{sec:system}.

\vspace{-0.1cm}
\section{System Design \& Implementation}
\label{sec:system}
\vspace{-0.1cm}
\subsection{System Overview}
\begin{figure}[t]
	\centering
	\includegraphics[width=0.96\linewidth]{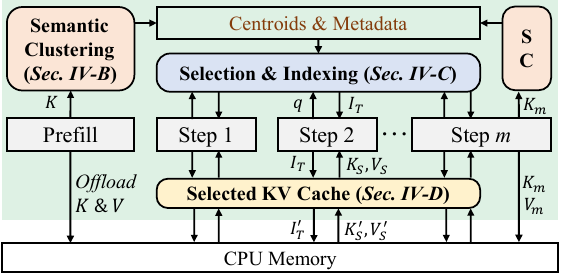}
	\caption{System overview of \SYSNAME{}. The green box represents components running on the GPU.
	 % \fix{Modify: 1. clustering and offloading in decoding steps; 2. add numbers in circle to the critical modules corresponding to subsections.}
     }
    \label{fig:overview}
    \vspace{-0.3cm}
\end{figure}

\begin{figure}[t]
	\centering
	\includegraphics[width=0.98\linewidth]{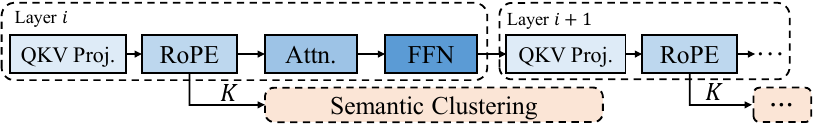}
	\caption{Overlapping of clustering with other operations.}
	\label{fig:cluster-overlap}
    \vspace{-0.5cm}
\end{figure}

% \vspace{-0.2cm}
The system overview is shown in Fig. \ref{fig:overview}.
During \textit{the prefill stage}, the key tensors are processed with semantic clustering (SC) on GPU, producing centroids and corresponding metadata. 
The generated KV tensors are offloaded to CPU memory. During \textit{the decoding stage}, as described in Section \ref{sec:algo-sel}, attention weights of the query vector $q$ and cluster centroids are computed to determine the importance of each cluster.
The results, along with clustering metadata, are used to generate indices (${I_T}$) for selected tokens, which are then used to load selected KV ($K_S,V_S$) from CPU memory to GPU memory for attention computation.
A cache for selected KV is maintained on GPU, so only KV not already cached ($K_S^\prime,V_S^\prime$)
% denoted as $K_S^\prime,V_S^\prime$ in Fig. \ref{fig:overview}, 
need to be loaded.
For every $m$ decoding steps, the clustering and KV offloading are performed for the $m$ generated tokens.

% In the following sections, we introduce how \SYSNAME{} achieves efficient clustering, selection and indexing in details.

\vspace{-0.1cm}
\subsection{Semantic Clustering}
% \vspace{-0.1cm}
\SYSNAME{} optimizes the efficiency of clustering at both system and kernel levels.

\textbf{System-level Optimization.}
% At the system level, 
As shown in Fig.~\ref{fig:cluster-overlap}, \SYSNAME{} applies clustering asynchronously, launching intermediately after the keys are computed from the QKV projection and RoPE modules. 
This allows clustering to be overlapped with other operations, including attention and FFN computation of the current layer, as well as the QKV projection and RoPE of the following layer. 
By overlapping these processes, \SYSNAME{} minimizes the overhead associated with clustering.

\begin{figure}[t]
	\centering
	\includegraphics[width=1\linewidth]{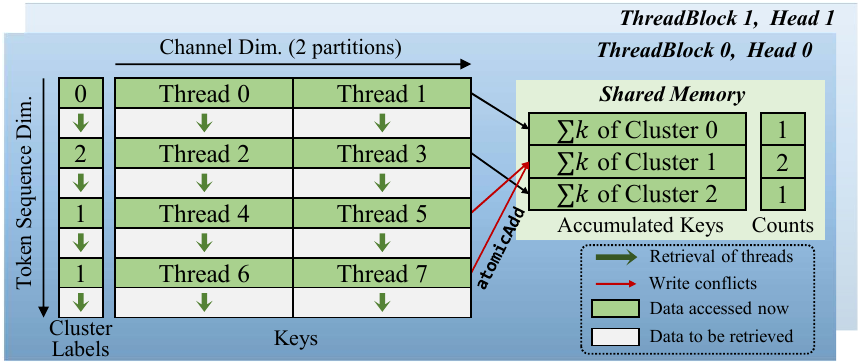}
	\caption{Execution of the centroid update kernel. 
    In this example, a \textit{ThreadBlock} consists of 8 threads ($\textit{BlockSize}=8$), and the channel dimension is divided into two partitions.}
	\label{fig:cluster-kernel}
\vspace{-0.6cm}
\end{figure}

\textbf{Kernel-level Optimization.}
% At the kernel level, 
Since clustering needs to be applied individually for each head, a key optimization is to achieve batched clustering across heads.
For the assignment step, which primarily involves \texttt{argmin} operations and matrix multiplications between keys and centroids, efficient batched Torch kernels are available \cite{torch2}. Therefore, we focus on optimizing the centroid update step and implement a custom CUDA kernel where different heads are processed in parallel by individual \textit{ThreadBlocks}, as shown in Fig. \ref{fig:cluster-kernel}.
The kernel retrieves keys, accumulates keys assigned to the same cluster, records corresponding accumulation counts in the shared memory, and computes the means as new centroids.
% after retrieval.

A potential issue 
% with the centroid update kernel 
is write conflict in shared memory, as accumulated keys and counts of the same cluster need to be computed using \texttt{atomicAdd}. To mitigate this, we apply several optimizations.
As shown in Fig. \ref{fig:cluster-kernel}, since distant tokens tend to be assigned to different clusters, we arrange threads in a strided manner along the sequence dimension to minimize the occurrence of concurrent processing of keys within the same cluster. 
Moreover, we partition the channel dimension into $P$ partitions, with each partition processed by one thread. This results in $\frac{BlockSize}{P}$ keys being processed in parallel.
There is a trade-off when choosing $P$: using a larger $P$ reduces the number of keys processed in parallel and increases the number of iterations along the sequence dimension, while using a smaller $P$ increases the number of iterations along the channel dimension within a thread, and potentially increases the occurrence of write conflicts.
To determine the optimal value of $P$, we perform offline profiling of different $P$ values. We find that for $BlockSize=512$ using $P=16$ or $P=32$ for 128 channels yields optimal performance.

\vspace{-0.1cm}
\subsection{Selection and Indexing}
% \vspace{-0.1cm}
\begin{figure}[t]
	\centering
	\includegraphics[width=0.95\linewidth]{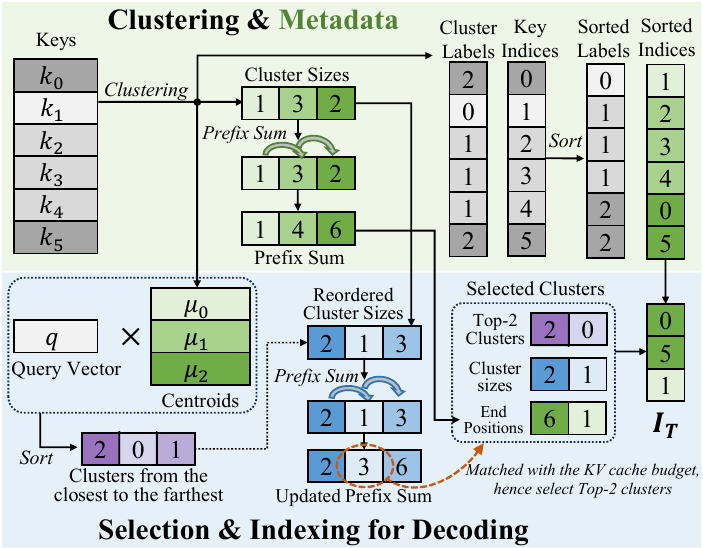}
	\caption{Process of selection and indexing. Green boxes represent stored metadata from clustering. Here, KV budget is 3.
	% , while the grey boxes indicate intermediates.
    }
	\label{fig:indexing}
    \vspace{-0.5cm}
\end{figure}

The details 
% of selection and indexing 
are shown in Fig. \ref{fig:indexing}.
After clustering, 
\SYSNAME{} stores cluster centroids and corresponding metadata, including cluster sizes, prefix sum, and sorted indices.
In this example, $k_0$ and $k_5$ are assigned to cluster 2, $k_1$ is assigned to cluster 0, and $k_2, k_3$ and $k_4$ are assigned to cluster 1. The sizes of three clusters can then be obtained.
% Consequently, the sizes of cluster 0, 1 and 2 are 1, 3 and 2, respectively.
Cluster labels and key indices are sorted by labels, and the sorted indices are stored.

During decoding, \SYSNAME{} computes attention weights between the query vector $q$ and cluster centroids $\mu$, sorting the clusters in descending order of attention weights to determine the closest clusters to $q$.
% For the example in Fig. \ref{fig:indexing}, the closest clusters are cluster 2, cluster 0 and cluster 1 in that order.
Next, \SYSNAME{} gathers and reorders their corresponding cluster sizes, and compute prefix sum.
% of the closest clusters from the size metadata, and compute their prefix sum to determine the number of selected clusters, denoted as $C_S$, based on the KV cache budget $B$.
% For the example in Fig. \ref{fig:indexing}, the closest clusters have sizes of 2, 1, and 3, their prefix sum becomes 2, 3, and 6. 
If we set the KV cache budget to three, which matches the second prefix sum, top-2 closest clusters will be selected.
% Given a budget $B=3$, the closest $C_S=2$ clusters are selected.
Then the labels, sizes, and end positions of selected clusters are gathered to determine the indices of selected tokens, $I_T$. 
Note that for cases where the summation of selected cluster sizes exceeds the budget, \SYSNAME{} trims tokens from the last selected cluster to adhere to the budget limit.

% Next, the sizes of selected clusters are extracted from the first $C_S$ elements of sizes of closest clusters, while the ending positions of selection are gathered from the size prefix sum metadata. 
% Finally, the indices of selected tokens, $I_T$, are gathered from sorted indices metadata based on selection sizes and ending positions.
% In Fig. \ref{fig:indexing}, the selected sizes are 2 and 1, and the ending positions are 6 and 1, for clusters 2 and 0, respectively.
% Using the size (2) and ending position (6) for cluster 2, token indices \{0, 5\} are extracted. Similarly, token index \{2\} is obtained from cluster 0's size (1) and ending position (1).
% The final indices of selected tokens are $I_T=\{0, 5, 1\}$.

Similarly to clustering, 
\SYSNAME{} implements efficient CUDA kernels, processing the indexing of KV heads in parallel with individual \textit{ThreadBlocks}.
Metadata related to cluster sizes, which is accessed frequently, is stored in shared memory for improving performance.

\vspace{-0.1cm}
\subsection{Caching KV of Selected Tokens}
\SYSNAME{} maintains a cluster-granularity cache on GPU to store KV of selected tokens, reducing unnecessary data transfers from CPU memory to GPU memory and enhancing performance.
During the decoding stage, the cache retains the KV of selected tokens from the last $R$ decoding steps, along with the labels of corresponding selected clusters.
At the current decoding step, the labels of selected clusters are compared with the cluster labels reserved by the cache. Only the KV of clusters not present in the cache are loaded from CPU memory.
The cache introduces a trade-off between memory usage and caching effectiveness. In practice, we find that setting $R=1$, i.e., retaining only the KV states from the last decoding step, strikes a good balance.

% Discussion: Gathering trade-off

\vspace{-0.1cm}
\section{Evaluation}
\subsection{Experimental Setup}
We evaluate \SYSNAME{} on an NVIDIA Ada 6000 GPU.
To assess model accuracy and output quality, we conduct experiments on eight datasets from LongBench \cite{longbench}, including 2WikiMQA, TriviaQA, HotpotQA, MultiFieldQA, MuSiQue, NarrativeQA, Qasper and GovReport.
These datasets cover a variety of tasks, such as single-doc QA, multi-doc QA, few shot learning and summarization, with context lengths reaching up to 32k.
% \cite{2wikimqa,joshi2017triviaqa,yang2018hotpotqa,li2002learning,trivedi2022musique,kovcisky2018narrativeqa,qasper,govreport}.
Additionally, we evaluate \SYSNAME{} on the PG19 dataset for the language modeling task \cite{pg19}.
We use the state-of-the-art long-context GLM4-9B-Chat model for evaluation of model accuracy \cite{glm2024}, which supports a context window of up to 128k tokens.

We compare \SYSNAME{} with two state-of-the-art KV cache compression methods: Quest \cite{quest} and InfiniGen \cite{infinigen}.
Most configurations of these methods remain as in their original settings, such as $page\_size$ for Quest and the partial weight ratio and selection threshold for InfiniGen.
To align with settings of Quest which does not apply selection on the first two layers of the model, we also disable selection and use the full KV cache for the first two layers in \SYSNAME{} and InfiniGen, in both model accuracy and inference performance evaluations. 
As both Quest and InfiniGen only support efficient inference for Llama-architecture models, we use Llama-3.1-8B for evaluation of inference performance.

\vspace{-0.1cm}
\subsection{Model Accuracy}
\label{sec:eval-acc}

\begin{figure}[t]
	\centering
	\includegraphics[width=\linewidth]{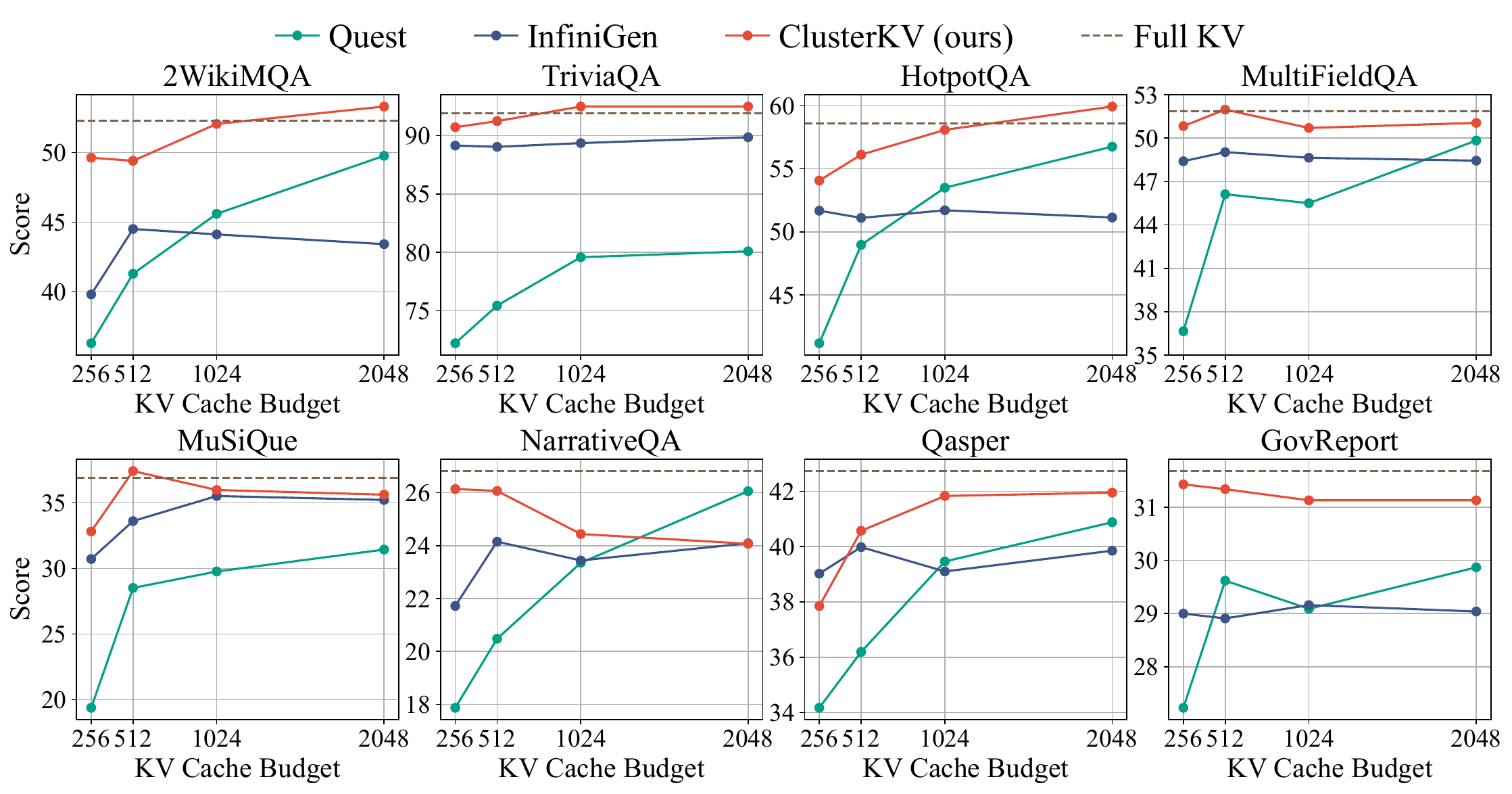}
	\caption{Results on LongBench of different methods.}
	\label{fig:longbench-tasks-res}
    \vspace{-0.3cm}
\end{figure}

\begin{table}[t]
\centering
\caption{Average scores on eight LongBench datasets.}
\begin{tabular}{|l|c|c|c|c|}
\hline
\diagbox{Method}{Budget} & 256 & 512 & 1024 & 2048 \\
\hline
Quest            & 35.63   & 40.83   & 43.23    & 45.59    \\
InfiniGen        & 43.69   & 45.04   & 45.13    & 45.14    \\
\textbf{ClusterKV (ours)} & \textbf{46.69}  & \textbf{48.02} & \textbf{48.34} & \textbf{48.7} \\
\hline
Full KV          & \multicolumn{4}{c|}{49.01}  \\
\hline
\end{tabular}
\label{tab:longbench-avg}
\vspace{-0.3cm}
\end{table}

% \begin{figure*}[t]
%     \centering
%     \begin{minipage}{0.6\linewidth}
%         \centering
%         \includegraphics[width=\linewidth]{exp_figs/longbench-glm4-32k.pdf}
%         \caption{Figure caption}
%         \label{fig:figure}
%     \end{minipage}
%     \hfill
%     \begin{minipage}{0.35\linewidth}
%         \centering
%         \begin{tabular}{|l|ccc|lll}
%         \cline{1-4}
%         Budget  & \multicolumn{1}{c|}{Quest} & \multicolumn{1}{c|}{InfiniGen} & ClusterKV &  &  &  \\ \cline{1-4}
%         256     & \multicolumn{1}{c|}{1}     & \multicolumn{1}{c|}{1}         & 1         &  &  &  \\ \cline{1-4}
%         512     & \multicolumn{1}{c|}{1}     & \multicolumn{1}{c|}{1}         & 1         &  &  &  \\ \cline{1-4}
%         1024    & \multicolumn{1}{c|}{1}     & \multicolumn{1}{c|}{1}         & 1         &  &  &  \\ \cline{1-4}
%         2048    & \multicolumn{1}{c|}{1}     & \multicolumn{1}{c|}{1}         & 1         &  &  &  \\ \cline{1-4}
%         Full KV & \multicolumn{3}{c|}{100}                                                &  &  &  \\ \cline{1-4}
%         \end{tabular}
%         \caption{Table caption}
%         \label{tab:table}
%     \end{minipage}
% \end{figure*}

\textbf{Results on LongBench.} Figure \ref{fig:longbench-tasks-res} presents the results on LongBench. ROUGE-L is used as the score metric for GovReport and F1 score is used for other tasks.
We evaluate methods under the KV cache budgets of 256, 512, 1024 and 2048 tokens. 
\SYSNAME{} outperforms Quest and InfiniGen in most settings, and achieves accuracy comparable to that of using the full KV cache with budgets as low as 1k to 2k tokens. 

The average scores across the eight tasks under different budgets are summarized in Table \ref{tab:longbench-avg}, where \SYSNAME{} demonstrates significant improvement over Quest and InfiniGen. 
% This is because InfiniGen selects \fix{XX} and Quest selects tokens on a page basis. Both of them overlook \fix{XX}. In contrast, \SYSNAME{} \fix{XX} group tokens based on their distances in semantic space and the cluster-based selection  
% Moreover, InfiniGen requires additional GPU memory space to store their partial query and keys.

% a more significant improvement over Quest than InfiniGen.
% This is because InfiniGen performs selection at the token granularity, while Quest selects pages with 16 continuous tokens.
% However, while InfiniGen reduces selection cost by using partial query and keys with channel dimensions reduced to 30\%, this demands additional GPU memory space. 
% Furthermore, the computation cost of per-token selection in InfiniGen results in inference latency that is comparable to directly using the full KV cache, as detailed in Section \ref{sec:eval-perf}.

\begin{figure}[t]
	\centering
	\includegraphics[width=0.7\linewidth]{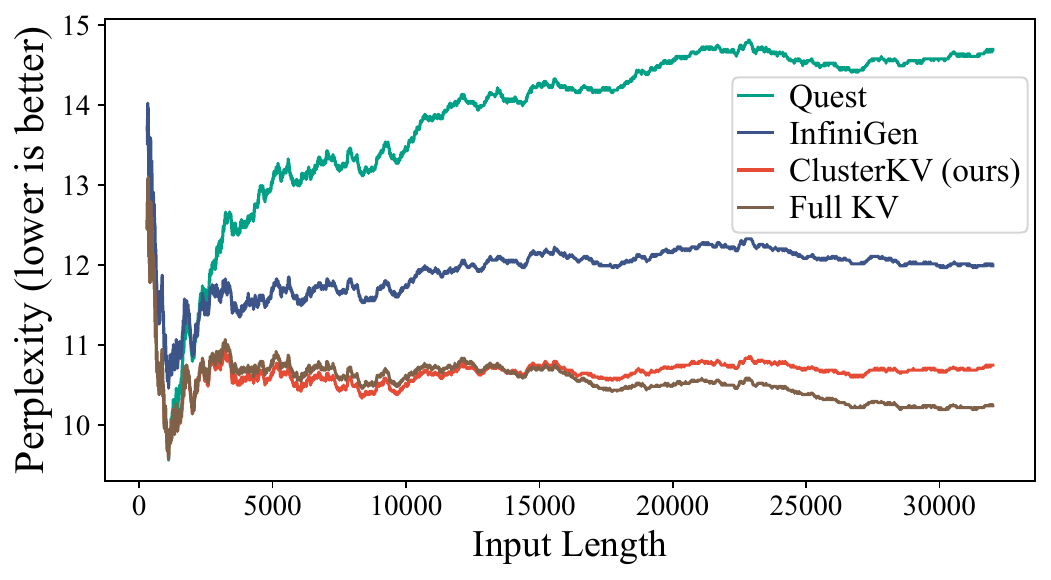}
	\caption{Perplexity of language modeling with a KV cache budget of 1024 tokens.}
	\label{fig:perplexity}
    \vspace{-0.3cm}
\end{figure}

\textbf{Results on Language Modeling.}
We report the perplexity of the language modeling task in Fig. \ref{fig:perplexity}. 
The prompts are from the PG19 test set, with input lengths ranging from 1 to 32000 tokens.
The KV cache budget is uniformly set to 1024 for \SYSNAME{}, Quest, and InfiniGen.
As shown, \SYSNAME{} closely aligns with the full KV, with a perplexity deviation up to 0.5, while Quest and InfiniGen show deviations of approximately 4 and 2, respectively.

\begin{figure}[t]
	\centering
	\includegraphics[width=1\linewidth]{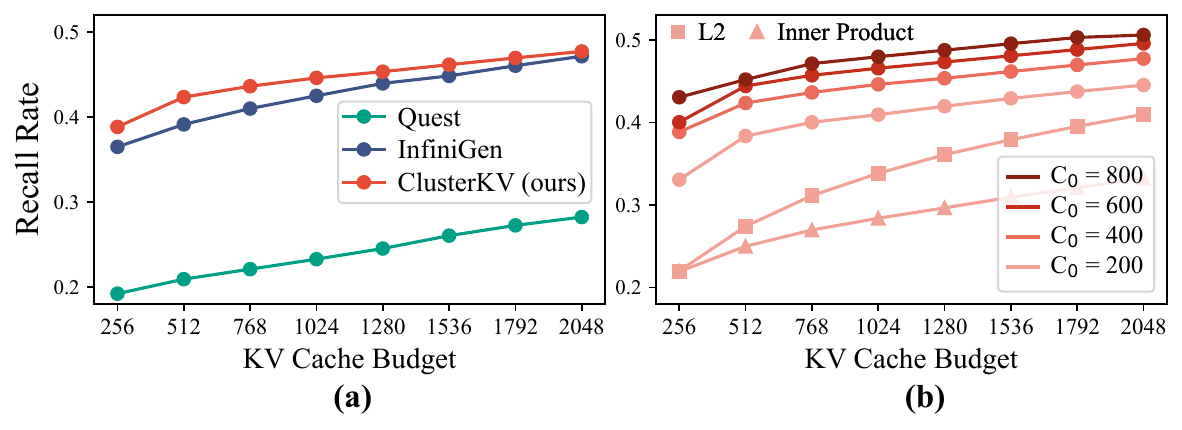}
	\caption{Recall rate of important tokens: (a) comparison between different methods, and (b) comparison between different configurations of \SYSNAME{}.}
	\label{fig:topk-recall}
    \vspace{-0.3cm}
\end{figure}

\textbf{Recall Rate of important tokens.}
We extract a sample from the NarrativeQA dataset with a context length of 32k and compute the recall rates of important tokens during inference for different methods.
The recall rate is defined as $\frac{|I_T\cap I^{true}_T|}{|I^{true}_T|}$, where $|I_T|=|I^{true}_T|=B$, $I_T$ represents the indices of selected tokens, and $I^{true}_T$ denotes the indices of tokens with the top-$B$ largest attention weights.
We report the recall rates averaged across layers, heads and decoding steps, and the budget $B$ is varied from 256 to 2048 in increments of 256.

As shown in Fig. \ref{fig:topk-recall}a, \SYSNAME{} achieves higher recall rates 
% over Quest and InfiniGen 
across all budgets.
We further explore the impact of different configurations for \SYSNAME{}, including clustering distance metrics and the number of clusters $C_0$.
In Fig. \ref{fig:topk-recall}b, cosine similarity used by \SYSNAME{} outperforms both L2 and inner product distance.
In addition, increasing $C_0$ improves recall rates, while the improvements become less significant when $C_0>400$. 
Consequently, we use $C_0=400=\frac{L}{80}$ as a balanced choice for accuracy and efficiency.

\vspace{-0.11cm}
\subsection{Inference Efficiency}
\label{sec:eval-perf}

\begin{figure}[t]
	\centering
	\includegraphics[width=1\linewidth]{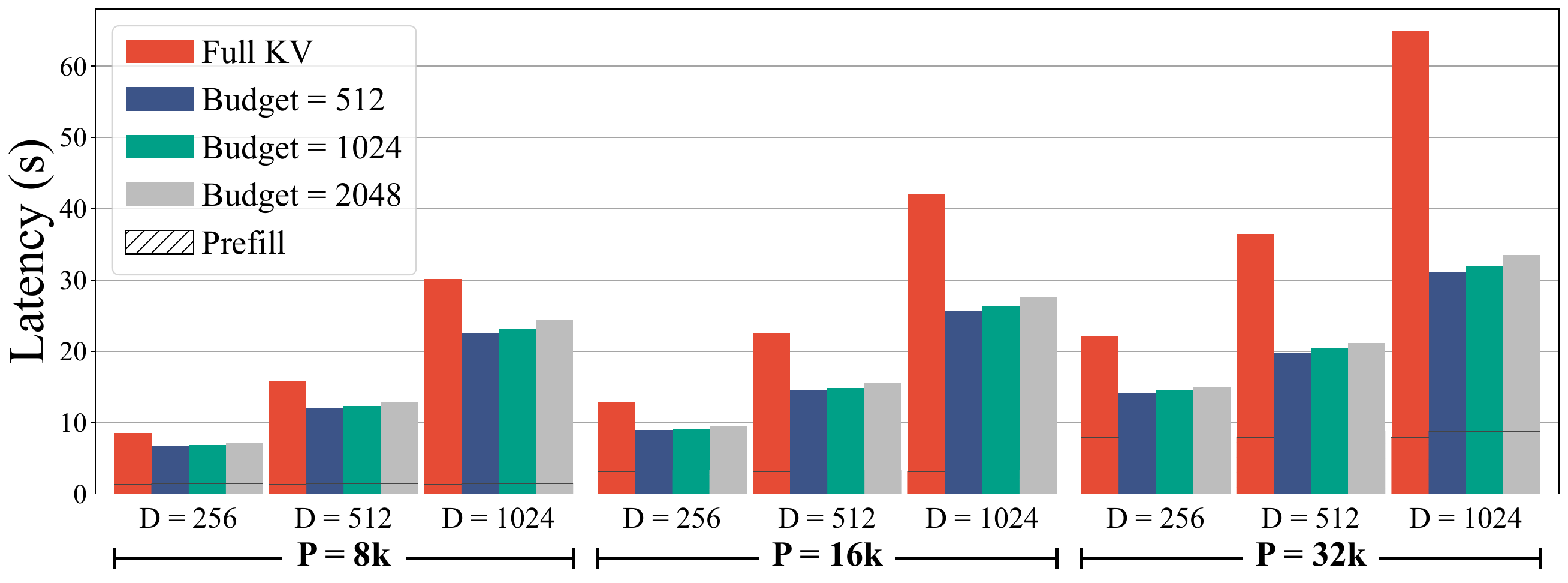}
	\caption{Comparison of inference latency between \SYSNAME{} and the full KV cache configuration.}
	\label{fig:perf-full}
    \vspace{-0.3cm}
\end{figure}

\textbf{Comparison with inference using full KV cache.}
We evaluate the inference latency of \SYSNAME{} under various KV cache budgets and compare it against the full KV cache configuration. 
The experiments are conducted with prompt lengths ($P$) ranging from 8k to 32k and decoding lengths ($D$) ranging from 256 to 1024.
As shown in Fig. \ref{fig:perf-full}, \SYSNAME{} achieves significant efficiency improvements. 
For $P=32k$ and $D=1024$, the speedup in latency is 2$\times$ with a budget of 1024 tokens, respectively. 
Additionally, the decoding throughput increases by up to 2.5$\times$.
We also analyze the time spent during prefill. As shown, the clustering overhead of \SYSNAME{} is minimal, accounting for only 6\% to 8\% of prefill and less than 2\% of the total inference time.

\begin{figure}[t]
	\centering
	\includegraphics[width=1\linewidth]{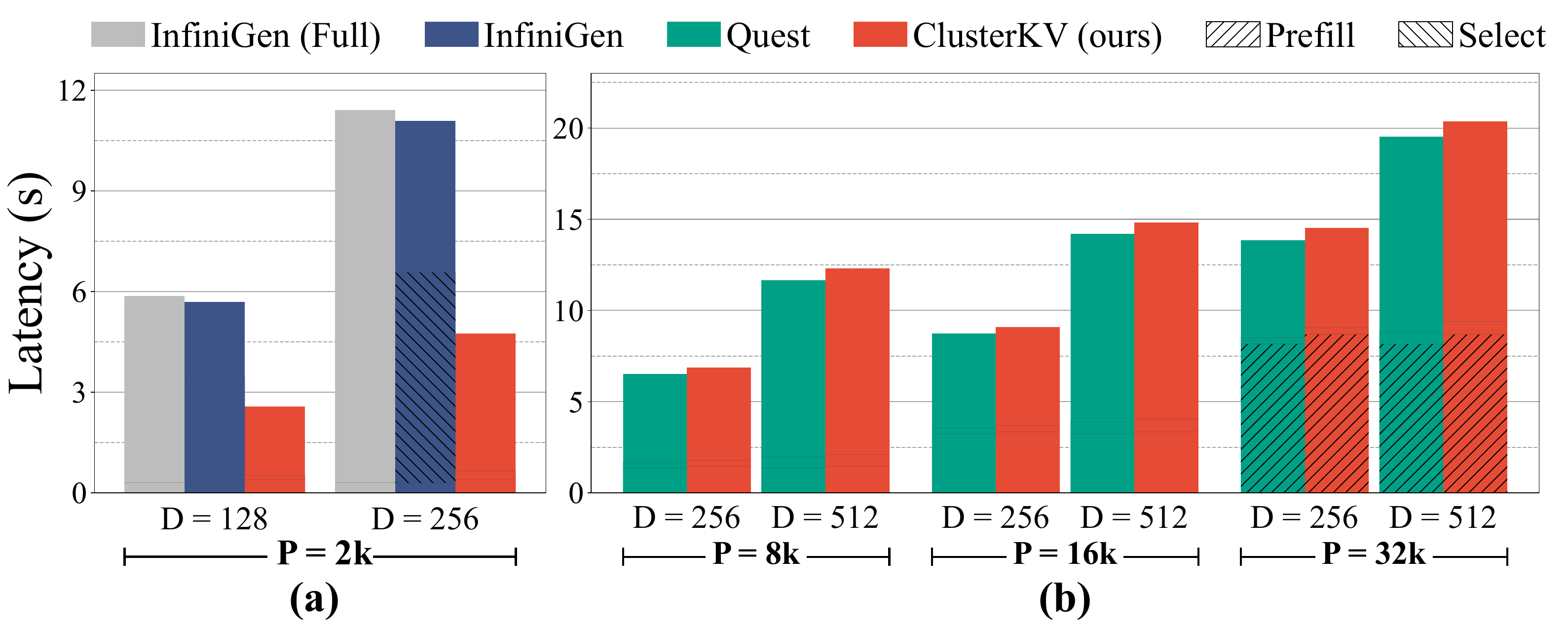}
	\caption{Comparison of inference latency: (a) \SYSNAME{} vs. InfiniGen with a budget of 256 tokens, and (b) \SYSNAME{} vs. Quest with a budget of 1k tokens.}
	\label{fig:perf}
    \vspace{-0.3cm}
\end{figure}

\textbf{Comparison with SoTA compression methods.}
InfiniGen is implemented based on FlexGen~\cite{flexgen}, which exclusively supports OPT \cite{zhang2022optopenpretrainedtransformer} models with a 2k context window from scratch and is hard to adapt to other models, so we compare \SYSNAME{} with InfiniGen using OPT-6.7B, as shown in Fig.~\ref{fig:perf}a. \SYSNAME{} achieves an average speedup of 2.3$\times$.
% under $P=2k$ the budget of 256 tokens, \SYSNAME{} achieves $\times$ and $\times$ speedups compared to InfiniGen for $D=128$ and $D=256$, respectively.
InfiniGen's latency is comparable to that of inference using the full KV cache, due to the high computational cost of its per-token selection, while \SYSNAME{}'s selection overhead accounts for only about 5\% of the total decoding latency.

We present the comparison between \SYSNAME{} and Quest in Fig. \ref{fig:perf}b, using Llama-3.1-8B model with a budget of 1k tokens.
As shown, \SYSNAME{} achieves performance very close to Quest, with latency deviations up to 5\%, while \SYSNAME{} delivers significantly higher model accuracy.

\textbf{Effectiveness of caching.}
We analyze the hit rates of our cluster-granularity cache during inference, using a 32k-tokens sample from the NarrativeQA dataset.
The average hit rates are 63\% and 74\% for $R=1$ and $R=2$, respectively. 
Compared to directly loading from CPU memory, the caching mechanism improves decoding throughput by 2.3$\times$ and 3$\times$, respectively.

% \vspace{-0.1cm}
\section{Conclusion}
% \vspace{-0.1cm}
This paper introduces \SYSNAME{}, achieving efficient and accurate KV cache compression, recalling tokens at the granularity of semantic clusters. 
\SYSNAME{} preserves model accuracy while significantly enhancing inference efficiency.

% \vspace{-0.1cm}
\section{Acknowledgment}
% \vspace{-0.1cm}
This work is sponsored by the National Natural Science Foundation of China (62472273, 62232015).

% \cite{llmlingua, streamingllm, h2o, keyformer, snapkv, pyramidkv, pyramidinfer, yoco-msra, cla-mit, infinigen, mqa, gqa,  minference, razor-attn, adakv, atom, kivi, intactkv, quest, loki,
% orca, vllm, distserve, splitwise, memserve, mooncake, sarathi, cached-attn, promptcache, cacheblend,
% flash-attn2, ring-attn
% } 

% \clearpage
\bibliographystyle{IEEEtran}
\bibliography{ref}

\end{document}